%% file: ecml_conf.tex
\documentclass[runningheads,a4paper]{llncs}
\usepackage{amssymb}
\usepackage{adjustbox}
\usepackage{caption}
\usepackage{subcaption}
\usepackage{pgfplots}
\usetikzlibrary{positioning,arrows}
\usetikzlibrary{shapes,decorations.pathmorphing, patterns}
\usepackage[algoruled]{algorithm2e}
\usepackage{graphicx}
\newtheorem{theo}{Theorem}

\newtheorem{assumption}{Assumption}

\usepackage{wrapfig}

\newcommand{\SGO}{\texttt{SGD}}

\newcommand{\ASGD}{\texttt{ASGD}}
\newcommand{\DSGD}{\texttt{DSGD}}

\newcommand{\Webspam}{\texttt{Webspam}}
\newcommand{\Epsilon}{\texttt{Epsilon}}

\newcommand{\AGDBC}{\texttt{ADG}_{\texttt{BC}}}
\newcommand{\AGDMF}{{\texttt{ADG}_{\texttt{MF}}}}
\newcommand{\SGrad}{\texttt{Sync-SVRG}}
\newcommand{\ASGrad}{\texttt{Async-SVRG}}

\newcommand{\RCV}{\texttt{RCV1}}
\newcommand{\loss}{\mathcal L}
\newcommand{\Insloss}{\ell} 
\newcommand{\Trn}{S}
\newcommand{\obs}{\mathbf{x}}

\newcommand{\X}{\textbf{w}}
\newcommand{\Y}{\textbf{v}}

\usepackage{url}
\usepackage{amsmath}
\urldef{\mailsa}\path|{bikash.joshi, franck.iutzeler, massih-reza.amini}@imag.fr|

\begin{document}

\mainmatter  

\title{An Asynchronous Distributed Framework for Large-scale Learning Based on Parameter Exchanges}


%
%
\author{Bikash Joshi, Franck Iutzeler, Massih-Reza Amini}%
%

\institute{University of Grenoble Alpes,\\
Grenoble, France\\
\mailsa\\}

%
%

\toctitle{Lecture Notes in Computer Science}
\tocauthor{Authors' Instructions}
\maketitle

\begin{abstract}
In many distributed learning  problems, the heterogeneous loading of computing machines may harm the overall performance of synchronous strategies. In this paper, we propose an effective asynchronous distributed framework for the minimization of a sum of smooth functions, where each machine performs iterations in parallel on its local function and updates a shared parameter asynchronously. In this way, all machines can continuously work even though they do not have the latest version of the shared parameter. We prove the convergence of the consistency of this general distributed asynchronous method for gradient iterations then show its efficiency on the matrix factorization problem for recommender systems and on binary classification. 
\end{abstract}

\input{introduction}

\input{async_strategy}
\input{Applications}

\input{Experiments}

\input{conclusion}

\bibliography{ecml}
\bibliographystyle{splncs03}
\end{document}

%% file: introduction.tex
\section{Introduction}

With the ever growing size of available data, distributed learning strategies where training sets are stored over $M$ connected machines have attracted much interest in both machine learning and optimization communities. In this paper, we propose a principal asynchronous way to minimize a general differentiable objective that can be written as:  
\begin{equation}
\label{eq:Opt}
\loss(\Y,\X)= \sum_{i=1}^M \loss_i(\Y_i,\X) 
\end{equation}
where $\Y= (\Y_1,..,\Y_M)$. In this model, each loss function $\loss_i$  depends on $(i)$ a \emph{local} version of parameter $\Y$, i.e. $\Y_i$, that does not need to be exchanged across different machines, and $(ii)$ a \emph{shared} parameter $\X$ that has to be exchanged. 

This formulation covers two common situations. First, when each loss $\loss_i$, depends only on \emph{local} versions of parameter $\Y$, the learning problem reduces to, $\min_{\Y} \sum_{i=1}^M \loss_i(\Y_i)$, which is a totally parallel problem that can be solved locally on each machine in parallel \cite{Bertsekas:1989}. 

The other extreme is a more typical case where each loss $\loss_i$, depends only on the global shared parameter $\X$ and the learning problem in this case reduces to, $\mathop{\text{min}}_{\X} \sum_{i=1}^M \loss_i(\X)$. This kind of problem is extremely common in ML when one wants to find the best predictor from a dataset split in several batches. Many deterministic and stochastic synchronous distributed algorithms have been recently proposed to solve this problem \cite{sra2012,Ho2013,Mairal15}. In most of these methods, the next global parameter is computed using updates based on its current version. In terms of implementation, the shared parameter is sent from each machine to a master node and is then broadcasted back into the network after integrating (mostly averaging) its local copies. For these synchronous methods, the loading of machines plays a central role in the convergence time of the whole system and in the extreme case, the slowest machine may become a bottleneck. To overcome this shortcoming, recent studies have considered asynchronous framework for distributed optimization \cite{xu2014globally,Tsung15,zhang2015fast,huo2016asynchronous,dean2012large}. However, these approaches suffer from mainly two drawbacks. First, some of these approaches \cite{xu2014globally,Tsung15} are based on a fixed delay time for broadcasting the parameter and the automatic tuning of this hyperparameter has to take into account the dynamic load of computing nodes in a network, and is a tedious task. Whereas others \cite{zhang2015fast,huo2016asynchronous,dean2012large} rely on communicating gradients after each mini-batch update, So, if the size of dataset grows large the communication cost will become huge especially for a large number of workers.

In this paper, we propose a novel asynchronous distributed framework for the minimization of the objective \eqref{eq:Opt}. In this framework, each worker machine sends its updated parameter values to a master machine, or also referred as the server, after finishing an iteration over its own local subpart of the data, and, it immediately begins a new iteration using: either the updated parameter copy received from the master machine (if it had received it from master machine during previous iteration) or it continues with its local updated parameter (if no update was received). Whereas the master, aggregates the received updates with its own local update whenever it finishes its iteration and broadcasts the updated parameter to all machines. In this way, all the machines have an overall view over the complete data, and the communication cost is significantly minimized as compared to the methods which broadcast the gradients after every mini-batch iteration \cite{zhang2015fast,huo2016asynchronous,dean2012large}. Thus the proposed method is totally asynchronous (non-blocking) and overcomes the bottleneck of slower machines in the distributed framework leading to a much faster convergence. We provide a proof of convergence of the updates to a (local) minimum of the overall objective function, and empirically show the efficiency of the proposed approach on the matrix factorization problem for recommender systems on NetFlix and MovieLens 10M datasets, as well as large-scale classification.

In the following section, we describe the proposed asynchronous distributed strategy, and derive two algorithms for large-scale binary classification and matrix factorization presented in Section \ref{sec:Applis}. Finally, Section \ref{sec:Exps}  presents experimental results corresponding to these two applications respectively.

%% file: async_strategy.tex
\section{Asynchronous Distributed Strategy}
\label{sec:2}

In this section, we present our proposed asynchronous distributed approach by first describing the deduced learning strategy. We then provide a consistency justification in the form of a convergence proof.

\subsection{Description}

\begin{figure*}[t!]
\centering
 \begin{tabular}{c c}
\includegraphics[width=.47\textwidth]{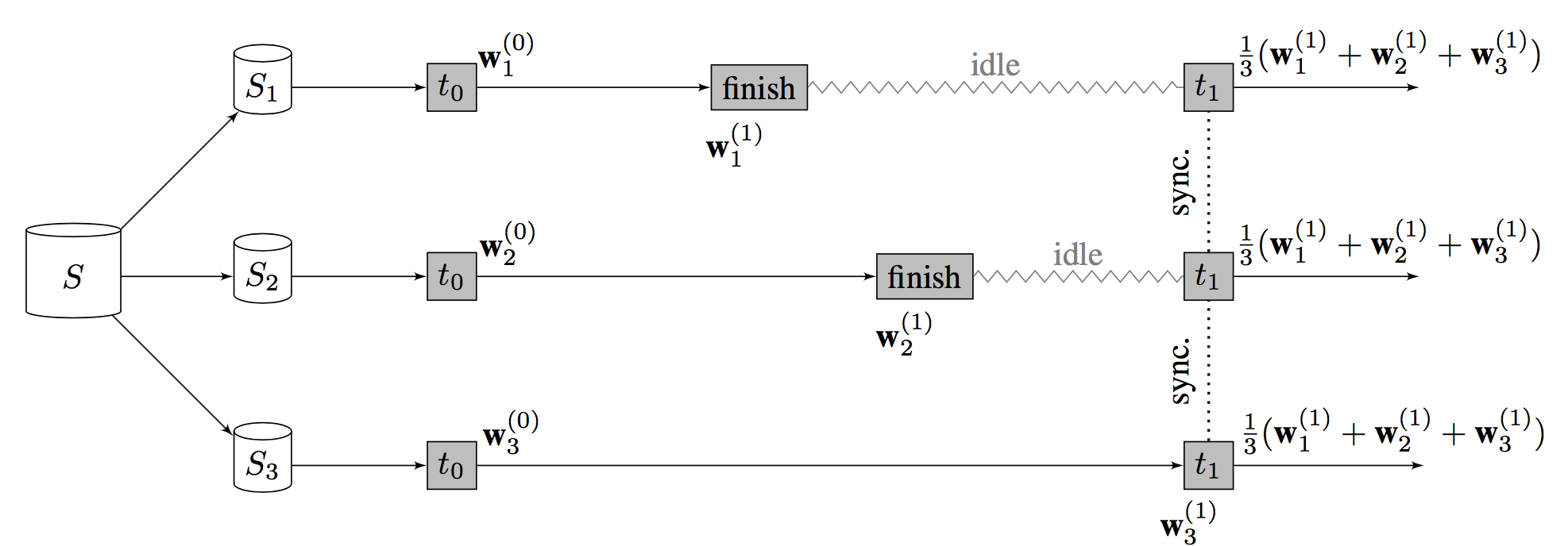}
&
\vspace{-1mm}\includegraphics[width=.51\textwidth]{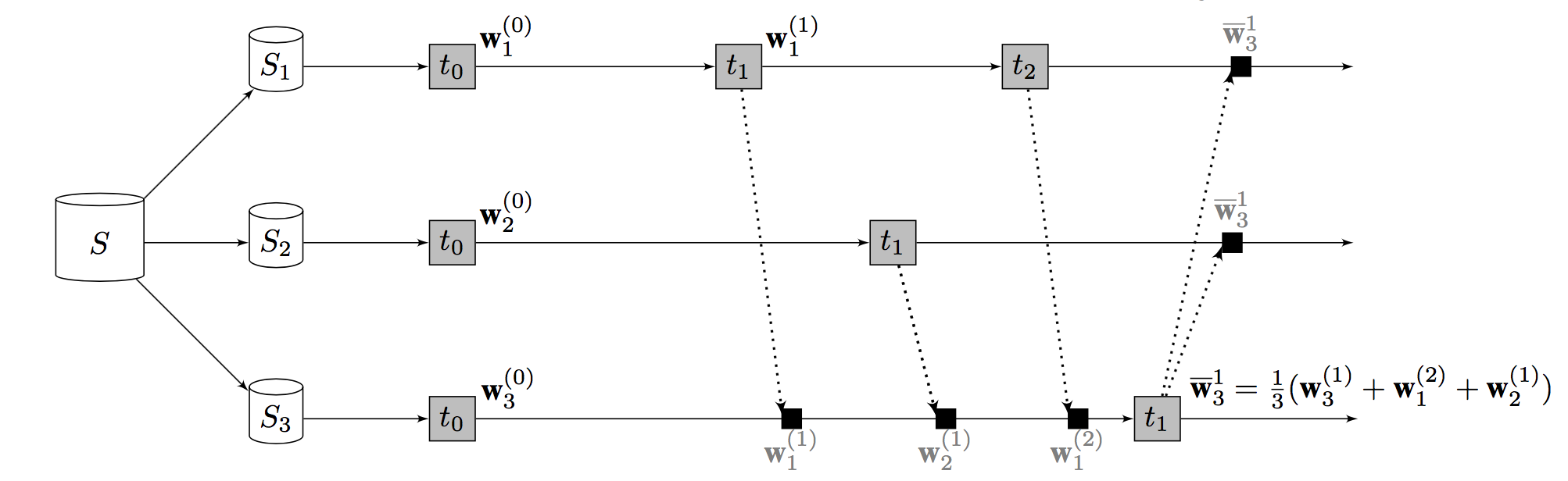} \\
(a) & (b) 

\end{tabular}
\caption{Diagrams of the distributed synchronous (a) and asynchronous (b) frameworks.}
\label{fig:sync}
\end{figure*}

The main challenge of distributed learning is to effectively partition the data into computing nodes, and efficiently perform communication between them. Indeed, in the synchronous case, the slowest node becomes the bottleneck of the whole system and a potentially large amount of computational time is lost (Figure~\ref{fig:sync} (a)).   

The main idea of our approach is that when a machine finishes an iteration over the subpart of the data it contains, it broadcasts its updated parameter values to the master node; which gathers the received parameter values from the workers (if any, and taking only the last one if multiple parameter values are received from one machine); and updates the parameter vector with the received updates. Then the updated parameter is broadcasted to worker nodes. In this way each computing node runs its iterations independently and gets rid of the synchronization bottleneck. Faster machines will perform their epochs faster, whereas the slower ones will be lagging on time but after finishing each epoch they will receive the most updated parameters from the master. This situation is depicted in Figure~\ref{fig:sync} (b).

The main difference with other distributed asynchronous algorithms proposed in the literature \cite{zhang2015fast,huo2016asynchronous}, our approach does not exchange gradients but rather parameter values updated after one complete pass over local subpart of the data. Although these quantities have the same sizes, the broadcasting of parameters performs better in practice, since they are exchanged after each epoch, whereas gradients need to be exchanged after every mini-batch update. 

\subsection{Consistency justification}
\label{sec:pb2}

In the case where the training data is partitioned  into $M$ batches $\{\mathcal{S}_1,\ldots,\mathcal{S}_M\}$, one for each computing machine, in the \textit{shared parameter} case, the objective Eq.~\eqref{eq:Opt} can be rewritten as 
\begin{equation}
\label{eq:DistributedSharedObj}
\loss(\X) =  \sum_{i=1}^M \loss_i(\X).
\end{equation}
Here we may take advantage of the differentiability of  $(\loss_i)_{i=1}^M$ and use a gradient algorithm to find a minimizer of the global objective, $\loss$. With a fixed stepsize gradient as an elementary operation before exchanging, we make the following assumptions :

\begin{assumption}[on the functions]
\label{hyp:f2} 
~~ 
\begin{itemize}
\item[a.] The objective function, $\loss$, has a unique minimizer $\X^\star$;
\item[b.] Each $\loss_i$ is differentiable and $\nabla \loss_i$ is $\frac{1}{L}$-cococercive, that is $\forall \X,\X' \in\mathbb{R}^d$:
\end{itemize}
\[
\langle \X - \X' ; \nabla \loss_i (\X) - \nabla \loss_i(\X') \rangle \geq \frac{1}{L} \|\nabla \loss_i (\X) - \nabla \loss_i(\X')  \|^2.
\]

\end{assumption}

As a consequence of the Baillon-Haddad theorem (Th.~18.15 in \cite{bauschke2011convex});  Assumption~\ref{hyp:f2} $(b)$ is notably verified whenever all functions $\loss_i$ are  convex and $L_i$-smooth,  that is differentiable with an $L_i$-Lipschitz continuous gradient with $L=\max_i L_i$. Also, if a function $\loss_i$ is $L_i$-smooth but not necessarily convex, then, considering $g_i= \loss_i+\lambda/2\|\cdot\|^2$, it comes that $\nabla g_i$ is $1/(2\lambda)$ cocoercive for $\lambda>L$ (see Prop.~2 in \cite{zhu1996co}). In our case, this means that if the (smooth) cost function is non-convex, then one can add a $\ell_2$ regularization term so that the sum function verifies the sought property. 

In Assumption~\ref{hyp:a}, we also make the rather mild assumption that the delays are bounded, meaning that no machine is infinitely slower than the others. More precisely, we consider that
the duration of its computation is bounded by $D$ in the sense that if machine $i$ finishes its computation at time $k+1$, then the value of the averaged parameter it used is at most $D$ ticks old. Mathematically, denoting the computation delay for machine $i$ at time $k$ by $d_i^k$, our bounded delay assumptions means that when machine $i$ finishes, say at time $k$, the (outdated) value of the averaged parameter it used is $\overline{\X}^{k-d_i^k}$ with $d_i^k\leq D$.

\begin{assumption}[on the algorithm] \label{hyp:a}
The delays are uniformly bounded, i.e. there is $D<\infty$ such that for any machine $i$ and iteration $k$; the delay $d_i^k \leq D$. 
\end{assumption}


\begin{wrapfigure}[10]{r}{0.71\textwidth} 
\textbf{Asynchronous Distributed Gradient update rule}\hrulefill\\
When machine $i$ finishes computing $\nabla \loss_i(\overline \X^{k-d_i^k} )$\\[0.1cm]
\hspace*{0.1cm} (\textbf{Local step}) at $i$:~  $\X_i^{k+1} = \overline{\X}^{k-d_i^k} - \gamma \nabla \loss_i( \overline{\X}^{k-d_i^k} ) \label{eq:UpdateRuleSGD}$\\[0.2cm]
\hspace*{0.1cm} (\textbf{Master step}) $\overline{\X}^{k+1} = \frac{1}{M} \sum_{j=1}^M  \X_j^{k+1}$\\[0.2cm]
\hspace*{0.1cm} \hphantom{(\textbf{Master step})} Broadcast $\overline{\X}^{k+1}$\\
\hrule
\end{wrapfigure}

The proposed Asynchronous Distributed update rule, corresponding to Figure~\ref{fig:sync} (b), is summarized in the  pseudo-code in the right. In the \textbf{local step}, all machines including the master update their parameters; and in the \textbf{master step}, once the master finishes its update, it broadcasts the aggregated parameters (from the latest received ones) to all workers. Furthermore,  using a gradient step as an elementary operation, the convergence of the algorithm can be proven with the attractive properties that the considered stepsizes can be chosen fixed, as in the standard gradient algorithm, and thus do not decay or depend on the delay; and that no assumptions are made on the distribution of the delays.

\begin{theo}[Convergence]
\label{th:cv2}
Suppose that Assumptions~\ref{hyp:f2} and \ref{hyp:a} hold. Let $\gamma\in]0,2/L[$. Then the sequence \emph{$(\overline{\X}^k)_k$} produced by our \emph{Asynchronous Distributed Gradient update rule} converges to $\X^\star$.
\end{theo}

\emph{Proof.} From Assumption~\ref{hyp:f2} $(i)$, $\X^\star$ is the unique minimizer of $\loss$ and $\nabla \loss(\X^\star) = \sum_{i=1}^M\nabla \loss_i(\X^\star) = 0$. Let us define for all $i=1,..,M$ $ \X_i^\star = \X^\star - \gamma \nabla \loss_i(\X^\star)$. Then at time $k$ for the updating machine $i$, it comes from the cocoercivity of $\nabla \loss_i$, Assumption 1 $(b)$; and the definition $\X_i^{k+1} = \overline{\X}^{k-d_i^k} - \gamma \nabla \loss_i( \overline{\X}^{k-d_i^k} )$:
\begin{align*}
 &\left\| \X_i^{k+1} - \X_i^\star \right\|^2   = \left\| \overline{\X}^{k-d_i^k}  - \gamma \nabla \loss_i( \overline{\X}^{k-d_i^k} ) -  ( \X^\star - \gamma \nabla \loss_i(\X^\star) ) \right\|^2 \\
&\leq \left\| \overline{\X}^{k-d_i^k}  -  \X^\star \right\|^2 + \gamma^2 \left\|  \nabla \loss_i( \overline{\X}^{k-d_i^k})  - \nabla \loss_i(\X^\star) \right\|^2  - \frac{2\gamma}{L}  \left\|  \nabla \loss_i( \overline{\X}^{k-d_i^k})  - \nabla \loss_i(\X^\star) \right\|^2.\end{align*}

Now by setting $\delta=\gamma\left( \frac{2}{L} - \gamma \right)>0$ we get:
\begin{align*}
\left\| \X_i^{k+1} - \X_i^\star \right\|^2  & \leq \left\| \overline{\X}^{k-d_i^k}  -  \X^\star \right\|^2   - \delta \left\|  \nabla \loss_i( \overline{\X}^{k-d_i^k})  - \nabla \loss_i(\X^\star) \right\|^2  \\
&= \!\!\left\| \frac{1}{M} \sum_{j=1}^M (\X_j^{k-d_i^k}  -  \X_j^\star) \right\|^2 \!\!\!\! - \!\delta \left\|  \nabla \loss_i( \overline{\X}^{k-d_i^k}) \! - \!\nabla \loss_i(\X^\star) \right\|^2  \\
&\leq \frac{1}{M} \sum_{j=1}^M \left\|  \X_j^{k-d_i^k} \!\!\! - \! \X_j^\star \right\|^2 - \delta \left\|  \nabla \loss_i( \overline{\X}^{k-d_i^k})  - \nabla \loss_i(\X^\star) \right\|^2,
\end{align*} 
where we used the fact that 
\[
\sum_{j=1}^M \X_j^\star = \sum_{j=1}^M \X^\star - \gamma \sum_{j=1}^M \nabla \loss_j(\X^\star) = M \X^\star.
\]
As the gradient of the objective $\nabla \loss(\X) = \sum_{j=1}^M \nabla \loss_j (\X)$ is null at $\X^\star$. The last inequality is due to the convexity of the squared norm. For all other $j\neq i$, $\left\|  \X_j^{k+1} -  \X_j^\star \right\|^2 =\left\|  \X_j^{k} -  \X_j^\star \right\|^2$.

Let $\mathbf{y}^{k}_{d}=( \left\|  \X_i^{k-d} -  \X_i^\star \right\|^2)_{i=1,..,M} $ be the size-$M$ vector of the individual errors at time $k-d$; and let  $\mathbf{y}^{k}$ be the size-$M(D+1)$ vector obtained by concatenating the $ (\mathbf{y}^{k}_{d})_{d=0,..,D}$. From $\mathbf{y}^{k}$  to $\mathbf{y}^{k+1}$, we have that i) the last $M$ values,  $\mathbf{y}^{k}_D$, are dropped as they cannot intervene as $D$ is the maximal delay; ii) the other ones are moved $M$ coordinates lower $ \mathbf{y}^{k+1}_{d+1} = \mathbf{y}^{k}_{d}$ for $d=0,..,D-1$; iii) for the first $M$ coordinates, they are copied from time $k$, $ \mathbf{y}^{k+1}_{0} = \mathbf{y}^{k}_{0}$, except for the $i$-th one which verifies  $\|  \X_i^{k+1} \!\!\! - \! \X_i^\star \|^2 \leq  \frac{1}{M} \sum_{j=1}^M \|  \X_j^{k-d_i^k} \!\!\! - \! \X_j^\star \|^2$ thus $ \mathbf{y}^{k+1}_{0} (i) \leq \frac{1}{M} \sum_{j=1}^M \mathbf{y}^{k}_{d_i^k} (j) $. Thus one can write  $\mathbf{y}^{k+1} \preceq A^{k+1}  \mathbf{y}^{k}$ where `$\preceq$' indicates the elementwise inequality and $A^{k+1}$ represents the linear (in)-equalities mentioned above. $A^{k+1}$, seen as a $(D+1)\times (D+1)$ block matrix has identities on its sub-diagonal, and the top left block is the identity except for line $i$ which has $1/M$ coefficients on the $M$ columns corresponding to $d_i^k$. One can notice that it is non-negative and the row sum is constant equal to $1$.

Taking the $\ell_\infty$-norm, we have  $\| \mathbf{y}^{k+1} \|_\infty \leq \| A^{k+1} \mathbf{y}^{k}\|_\infty \leq  \| A^{k+1} \|_\infty \| \mathbf{y}^{k}\|_\infty \leq \| \mathbf{y}^{k}\|_\infty $ as the $\ell_\infty$-induced matrix $\|~\cdot~\|_\infty$ is the maximal row sum and all rows of non-negative matrix $ A^{k+1} $ have unit sum.  This means that $( \| \mathbf{y}^{k}\|_\infty )_k $ is a converging sequence, say to some value $\alpha$. Now, suppose that there is some coordinate that is strictly lower than $\alpha$, then it cannot be equal to $\alpha$ or greater anymore due to the above inequality; this means, that as the communication time is bounded, any coordinate holding the value $\alpha$ will have to (strictly) decrease due to the averaging with the strictly lower coordinate, which contradicts $\alpha$ being the limit of sequence $( \| \mathbf{y}^{k}\|_\infty )_k $. Thus, all errors converge to the same value which means that $\|  \nabla \loss_i( \overline{\X}^{k-d_i^k})  - \nabla \loss_i(\X^\star) \|^2 \to 0$, implying that all $ \X_i^k$ and thus $\overline{\X}^k$ converge. Furthermore, all limits points of $\overline{\X}^k$ null the gradient of $\loss$; $\X^\star$ being unique (Assumption 1 $(i)$), the convergence ensues. \qed

\bigskip

One can notice that using this asynchronous framework, the machines local parameters all converge to different values while their sum converge to the sought minimizer. As this sum is received after each iteration, the agents also have individual knowledge of the full minimizer. 
Finally, the tools used in this proof make it adaptable to a wide range of elementary operations verifying cocoercive contraction properties. For instance, if the loss has a smooth and a non-smooth part, the gradient step can be replaced by a proximal gradient step. Other possible extensions here include the Alternating Direction Method of Multipliers (ADMM) and Primal-Dual algorithms.

%% file: Applications.tex
\section{Applications}
\label{sec:Applis}
In the following sections we present two algorithms for the estimation of parameters on each machine, corresponding to the \textbf{local step} of the proposed Asynchronous Distributed Gradient update rule, for large-scale binary classification (Section \ref{sec:BC}) and matrix factorization for recommender systems (Section \ref{sec:MF}). 

\subsection{Asynchronous Distributed Gradient for Binary Classification ($\AGDBC$)}
\label{sec:BC}

For the classification problem, we consider the following convex loss function~:
\begin{equation}
\label{eq:Opt}
\loss(\X)= \frac{1}{n} \sum_{i=1}^n \Insloss(\X,\obs_i,y_i),
\end{equation}
defined over a training set of size $n$,  $\Trn=\{(\obs_i,y_i); i\in\{1,\ldots,n\}\}\in (\mathbb R^d\times \{-1,+1\})^n$,  where  the instantaneous loss associated to example $(\obs_i,y_i)\in \Trn$, $\Insloss(\X,\obs_i,y_i)$ is the $\ell_2$-regularized logistic surrogate~:
\begin{equation}
\label{eq:l2log}
\Insloss(\X,\obs_i,y_i) = \log(1 + \exp(-y_{i}\X^\top\obs_i)) + \frac{\lambda}{2n}||\X||^{2},
\end{equation}
where $\lambda\geq 0$ is a regularization parameter. In order to have an accelerated update of the parameters on a given machine, we rely on a \emph{variance reduced} variant of the Stochastic Gradient Descent (\SGO) algorithm.  Different such variants proposed recently, like SVRG \cite{johnson2013accelerating} or SAG/SAGA \cite{roux2012stochastic,defazio2014saga} reduce the variance caused through random-sampling in \SGO\ by occasionally computing full-gradients. As a result, this reduction in variance contributes to better convergence properties when using fixed learning rates. 

The distributed memory algorithm, corresponding to the \textbf{local step} in a computing  machine $j\in\{1,\ldots,m\}$ is shown in Algorithm \ref{algo:async_svrg_worker}.
Let $\widetilde{\X}$ be the last received aggregated parameter from the master, or the last updated parameter estimated locally if the computation finished before a new aggregated parameter has been received. A local average gradient is then estimated using the local subpart of the data stored in machine $j$; $\widetilde{\mu}_j = \bar{\nabla} \loss_j(\widetilde{\X})$. Considering a mini-batch  $I^t_j$ at the inner iteration $t$ of the computing machine $j$, the current parameter $\X^t$ is then updated as:
\begin{equation}
    \X^{t+1} \leftarrow \X^{t} -    
    \frac{\gamma}{|I^{t}_j|} \sum_{(\obs_i,y_i)\in I^{t}_j} (\nabla \Insloss(\X^{t},\obs_i,y_i) - \nabla \Insloss(\widetilde{\X},\obs_i,y_i) + \widetilde{\mu}_j),
\end{equation}
where $\gamma$ is the learning rate. This modification in update rule of \SGO\ is similar to the one of SVRG  \cite{johnson2013accelerating} with the difference that the local average gradient here is computed over the aggregated parameter sent by the master using the local subpart of the data, rather than it would be estimated over the whole data as in SVRG. The rational of using this slightly different version of SVRG, is that in the standard case it has been shown that SVRG reduces the variance of the algorithm near the convergence point, and it has a linear convergence rate.

 Each machine performs parameter update on their local data and after each iteration the computing machines send the updated parameter to the master which directly responds by sending the averaged common parameter using the last gathered updates (\textbf{Master step}). In this way, all the machines have an overall view of the parameter updates from whole data, while only working with their local data.  



\begin{algorithm}[H]
\textbf{Input}: Maximum number of iterations $T$, batch size $B$ and learning rate $\gamma$ \\
\textbf{Initialize}: $*$~ \underline{Receive} parameter $\widetilde{\X}\in\mathbb{R}^{d}$ from the master, or use the last parameter estimation happened before a new reception \;
$*~{\X}^0\leftarrow \widetilde{\X}$\;
$*$~Compute $\tilde \mu_j\leftarrow\bar \nabla \loss_{j}(\widetilde{\X})$ \;
\For{$t=0,.., T-1 $}
    {
         Randomly pick a mini-batch $I_j^{t}$  of size $B$ in the subpart of the data stored in machine $j$\;
        \textbf{Update} $\X^{t+1} \leftarrow \X^{t} - \frac{\gamma}{|I^{t}_j|} \sum_{(\obs_i,y_i)\in I^{t}_j} (\nabla \Insloss(\X^{t},\obs_i,y_i) - \nabla \Insloss(\widetilde{\X},\obs_i,y_i) + \widetilde{\mu}_j)$\;
    }
    $\widetilde{\X}\leftarrow \X^{T}$ and \underline{send} $\X^T$ to the master.
\caption{$\AGDBC$, \textbf{local step} in the computing machine $j\in\{1,\ldots,m\}$}\label{algo:async_svrg_worker}
\end{algorithm}

\subsection{Asynchronous Distributed \SGO\ for Matrix Factorization ($\AGDMF$)}
\label{sec:MF}

The problem of matrix factorization for collaborative filtering captured much attention, especially after the Netflix prize \cite{koren2009matrix}. The premise behind this approach is to approximate a large rating matrix $R$ with the multiplication of two low-dimensional factor matrices $P$ and $Q$, i.e. $R\approx \hat{R} =  PQ^\top$ that model respectively users and items in the same latent space. For a pair of user and item $(u,i)$ for which a rating $r_{ui}$ exists, the corresponding instantaneous loss is defined as $\ell_2$-regularized quadratic error:
\begin{equation}
\label{eqn:SGD1}
 \Insloss(P,Q,u,i)= \left(r_{ui} - q_{i}^{\top}p_{u}\right)^2 + \lambda(|| p_{u} ||^{2} + || q_{i} ||^2 ),
\end{equation}
where $p_u$ (resp. $q_i$) is $u$-th line of $P$ (resp. $i$-th line of $Q$) and  $\lambda\geq 0$ is a regularization parameter. The global objective is hence~:
\begin{equation}
\label{eqn:pb}
\loss(P,Q)=\sum_{(u,i) : r_{ui} \text{exists}}  \Insloss(P,Q,u,i).
\end{equation}
Note that instantaneous error $ \Insloss(P,Q,u,i)$ depends only on $P$ and $Q$ through $p_{u}$ and $q_{i}$; however, item $i$ may also be rated by user $u'$ so that the optimal factor $q_{i}$ depends on both $p_{u}$ and $p_{u'}$.

For this problem, \SGO\ was found to offer a high prediction accuracy on different recommender system datasets. In this case, the approach proceeds as follows: at each iteration $k$, i) select a user/item pair  $(u^k,i^k)$  for which a rating exists; ii) perform a gradient step on $\Insloss(P,Q,u^k,i^k)$. Here stochasticity is used in the sense that the gradient on $ \Insloss(P,Q,u^k,i^k)$ can be seen as an approximation of the gradient on an underlying global model but the choice of the considered users/items may or may not be random depending on the algorithm.

Despite its simplicity, there are several computational challenges associated with this problem. 
As previously, performing {\SGO} sequentially on a single machine takes unacceptably
large amount of time to converge for common rating matrices of several million ratings.
So, there is a need to perform {\SGO} in an efficient distributed manner for such large datasets.
However, parallelizing {\SGO} is not trivial. A drawback 
of a straightforward implementation is that updates on
factor matrices might not be independent. For example, for
training points that lie on same rows (i.e. ratings corresponding to the same users), an {\SGO} step modifies the same corresponding rows in factor matrix $P$; thus, these points cannot be learnt over in parallel and efficient
communication between the computing nodes is necessary to synchronize the updates on factor matrices.

A popular approach in this case is to divide the rating matrix into
several blocks and run gradient on each of the blocks on
distinct machines. From the decomposition $\hat{R} =  PQ^\top$, one can see that if the rating matrix is divided by row-blocks, $\hat{R}_b =  P_bQ^\top$, that is; the block $b$ of $\hat{R}$ depends only  on the block $b$ of $P$ then, the block-split problem writes:
\begin{equation}
\label{eqn:pb-split}
\min_{P,Q} \sum_{\text{blocks }b}  \left[\sum_{(u,i) : r_{b_{ui}} \text{exists}}  \Insloss(P_b,Q,u,i)\right].
\end{equation}

Factor matrices are thus updated independently on each
machine for the corresponding ratings. Even though the
rating matrix parts on each machine are different, the factor
matrix updates are not independent. So, after each
epoch the factor matrices present in each machine are synchronized.
We refer to this approach as Synchronous {\SGO}, as all machines synchronize their updates after every epoch. One example of such algorithm is \texttt{ASGD} proposed in \cite{makari2015shared}.

Another popular approach, referred to as Distributed \SGO\ (\texttt{DSGD}) \cite{gemulla2011large}, divides the rating matrix into set of disjoint blocks with non-overlapping rows and columns. A set of such disjoint blocks is named stratum, and the number of stratums in the rating matrix is fixed to the number of machines to be used in parallel.  These mutually independent sub-blocks in a stratum are processed in parallel and the updated parameters are synchronized after each stratum is processed (i.e. a subepoch). So, this method requires several synchronizations within an epoch which may hurt the computational performance.  

The main challenge of these distributed approaches is to effectively partition the data into computing nodes, and efficiently perform communication between them. Indeed, in the situation above, the slowest node becomes the bottleneck of the whole system.

In order to apply the asynchronous distributed  strategy to this problem (referred to as {$\AGDMF$} in the following), we split the rating matrix in row-wise manner. In this case, we only need to communicate the matrices $Q$ between machines, whereas the matrices $P$ are updated locally, corresponding to each sub-part, and are later concatenated at the end of the operation. Due to the shared variable, the \textbf{local step} of the algorithm has to be slightly adapted as shown in Algorithm \ref{algo:async_mf_worker}. As previously, \textbf{the master step} remains the same.

\begin{algorithm}[H]
Parameters: learning rate $\gamma$ \\
Initialize: $P_j$  \\
\underline{Receive} matrix $Q$ from the master\;
From the subpart of the data stored in machine $j$, \textbf{pick} randomly $(u,i)$ for which $r_{ui}$ exists \;
$(P_j,Q_j) \leftarrow (P_j,Q) - \gamma \nabla \Insloss(P_j,Q,u,i)$\;
\underline{Send} $Q_j$ to the master\;
\caption{$\AGDMF$ \textbf{local step} in the computing machine $j\in\{1,\ldots,m\}$}\label{algo:async_mf_worker}
\end{algorithm}

%% file: Experiments.tex
\section{Experimental Results}
\label{sec:Exps}
We conducted a number of experiments aimed at testing the behaviour of the proposed $\AGDBC$ and $\AGDMF$ on large scale classification and matrix factorization for recommender systems by comparing them to the state-of-the-art distributed approaches

\subsection{Experimental Results for Binary Classification}
\label{sec:Exp_BC}
In the first set of experiments we study the convergence and the communication overhead of the proposed $\AGDBC$ algorithm. 



\noindent \textbf{Datasets:}
We performed our experiments on two popular large-scale binary classification datasets: \Epsilon~and \RCV\footnote{\url{https://www.csie.ntu.edu.tw/~cjlin/libsvmtools/datasets/}}. The various characteristics of the datasets are presented in Table \ref{tab:dataset_properties}.

\begin{table}[h!]
\caption{Characteristics of Datasets used in our experiments. }
\label{tab:dataset_properties}
\tabcolsep=0.001cm
\centering
  \begin{tabular}{ c c c c c }
  \hline
  Dataset & Training Size ~~ & Test Size ~~ & Feature Dimension ~~ & $\#nonzeros$ \\
  
  \hline
     \Epsilon~~ & 400000~~ & 100000~~& 2000 & $10^9$   \\
   \RCV  &  558112~~ &  139529~~ &  47236~~~ &  51,055,210  \\ 
  \hline
   
\end{tabular}
\end{table}

\noindent \textbf{Baselines:}
We compare our approach with the following methods which also consider totally distributed scenario without shared memory.

\begin{itemize}
     \item The proposed approach {$\AGDBC$} (Section \ref{sec:BC}),
    \item \SGrad, SVRG based method \cite{johnson2013accelerating} with synchronization of gradients after every mini-batch update. 
    \item \ASGrad: Distributed architecture proposed in \cite{huo2016asynchronous}, which asynchronously communicate gradients after every mini-batch updates.   
\end{itemize}

Since the asynchronous methods were quite sensitive to initial point, we performed a synchronized gradient step during the first pass over the data. This gave a stable start for all the algorithms. 

\smallskip

\noindent \textbf{Platform:} Experiments were conducted in a platform with 7 disparate servers without shared memory. The code was implemented using a python module \texttt{mpi4py} using OpenMPI\footnote{\url{https://www.open-mpi.org/}} as the MPI library. 

\smallskip

\noindent \textbf{Hyper-parameters:} In all the experiments, we used a fixed regularization rate, $\lambda = \frac{1}{n}$, where $n$ is the size of the initial training set. The fixed learning rates were chosen from a set of values in range $\{ 10^{-4}, 10^{-3}, 10^{-2}, 10^{-1}$\} and the reported performance were the best obtained with one of those stepsizes. The mini-batch size for \Epsilon\ and \RCV\ datasets were respectively fixed to $10$ and $20$.  

\smallskip

\noindent \textbf{Evaluation Measures:} Convergence result was evaluated in terms of minimization of objective function over time. The communication overhead incurred by each algorithm  in the network as well as the communication time are shown in terms of the total number of send/receive calls.

  \begin{figure}[t]
  \centering
\begin{tabular}{cc}
      \includegraphics[width=.5\textwidth]{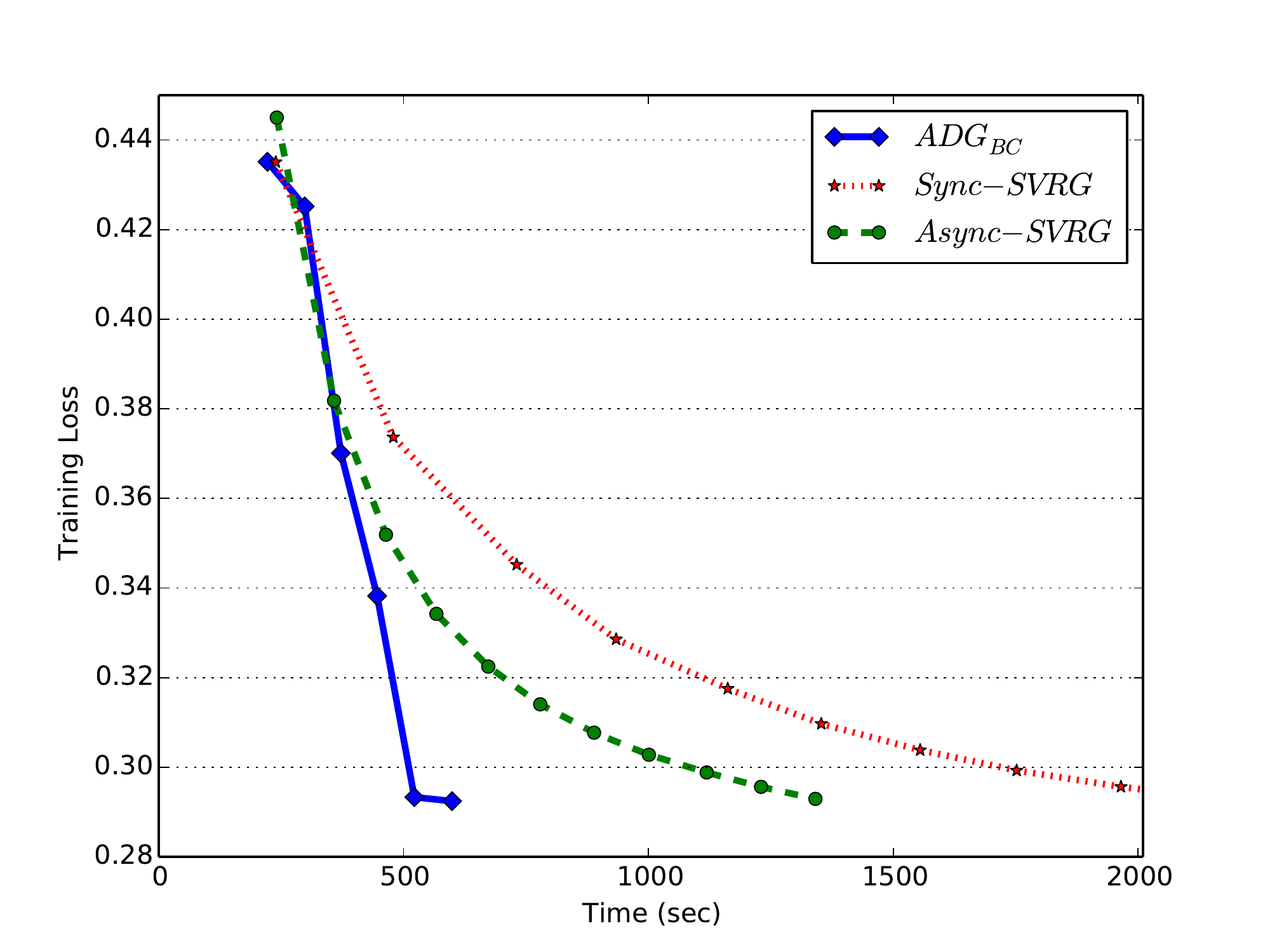} &
      \includegraphics[width=.5\textwidth]{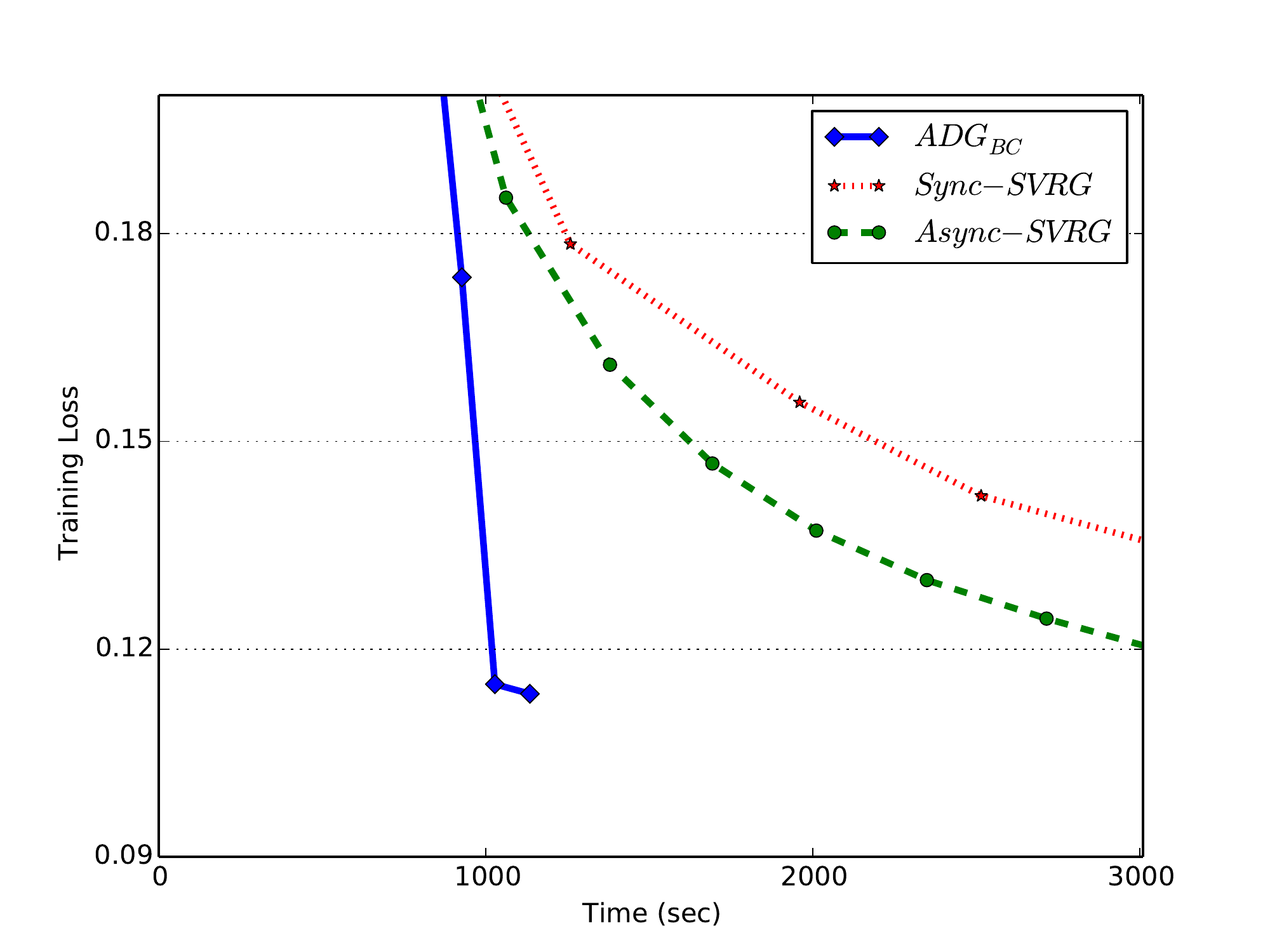}\\
      (a) \Epsilon\ dataset& (b) \RCV\ dataset 
\end{tabular}
  \caption{Training Loss Vs Time Plot for (a) \Epsilon\ and (b) \RCV\ Datasets}
  \label{fig:comparison_train_loss}
\end{figure}

\subsubsection{Evaluation of Convergence Time.}

 Figure $\ref{fig:comparison_train_loss}$ compares the convergence results for the three methods on all datasets. The convergence results are presented in terms of minimization of the objective function in the training sub-part of the data on the master machine. As It can be observed the proposed method $\AGDBC$ converges much faster than the other two methods. It can be seen that this behavior becomes more noticeable for larger datasets. For example on the \RCV\ collection, $\AGDBC$ converges three times faster than the other methods. Also it is to be noted that the difference in the convergence speed can become even larger if some of the machines are extremely overloaded, which is generally the case in the cluster environments.

\subsubsection{Communication Overhead}

\begin{table}[b]
\centering
\begin{tabular}{c cc c cc}
\hline
  &  \multicolumn{2}{c}{\Epsilon} & &  \multicolumn{2}{c}{\RCV} \\ \cline{2-3}\cline{5-6}

Methods  & Number of Calls ~~ & time (sec)  & & Number of Calls ~~  & time (sec) \\ \hline

$\SGrad$ &  108009 & 589.9 & &  83711 & 4756.25\\ 
$\ASGrad$ & 108000 & 110.03 & &  30701 & 733.47\\ 
\textbf{$\AGDBC$} &  \textbf{12004}& \textbf{29.6} & & \textbf{8380} & \textbf{631.92}\\ \hline

\end{tabular}
\caption{Comparison of the communication overhead for baselines}
\label{tab:comparison_comm}
\end{table}

\setcounter{figure}{2}
  \begin{figure}[t]
      \centering
      \begin{tabular}{cc}
      \includegraphics[width=.5\textwidth]{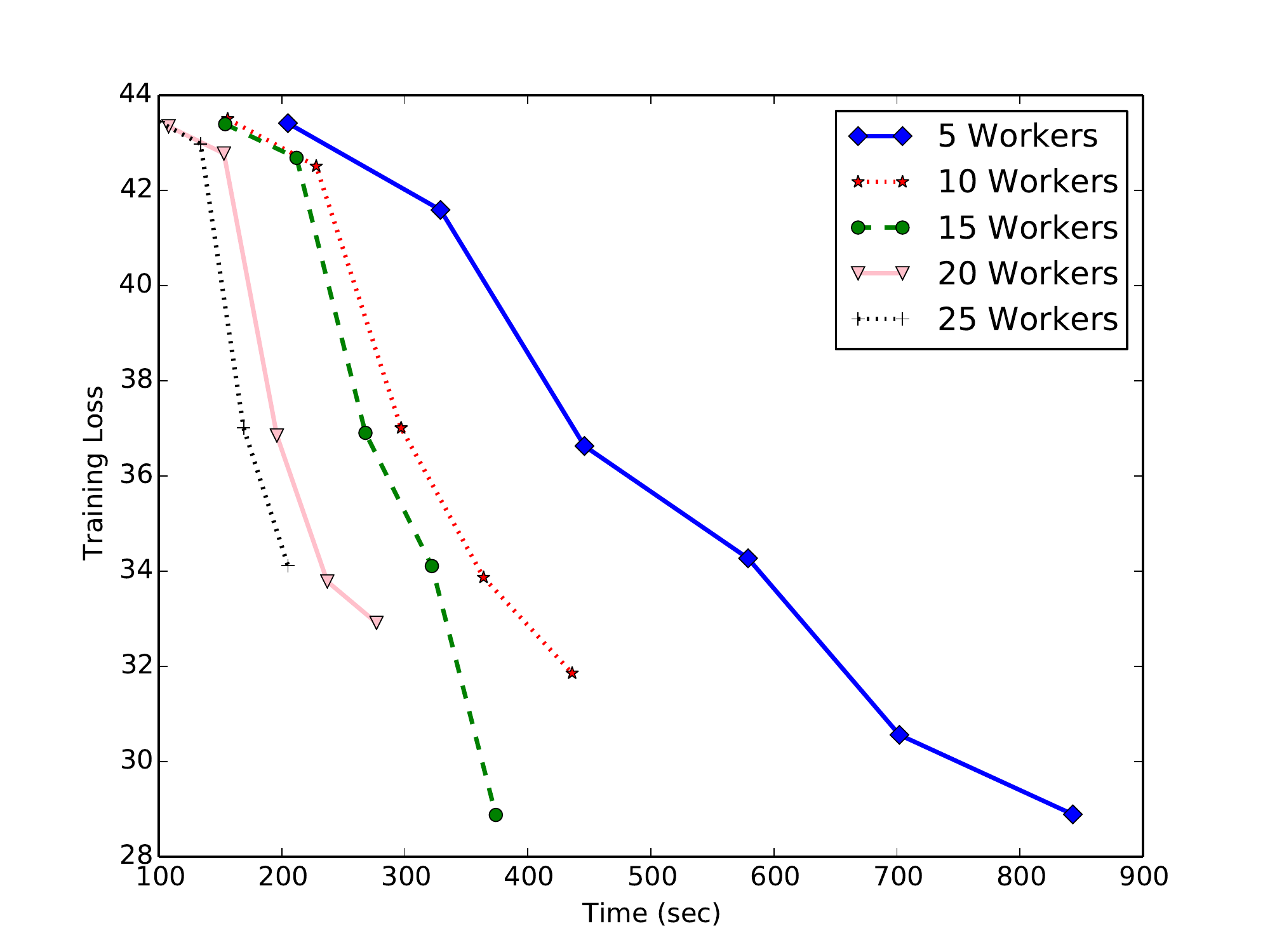} &  \includegraphics[width=.5\textwidth]{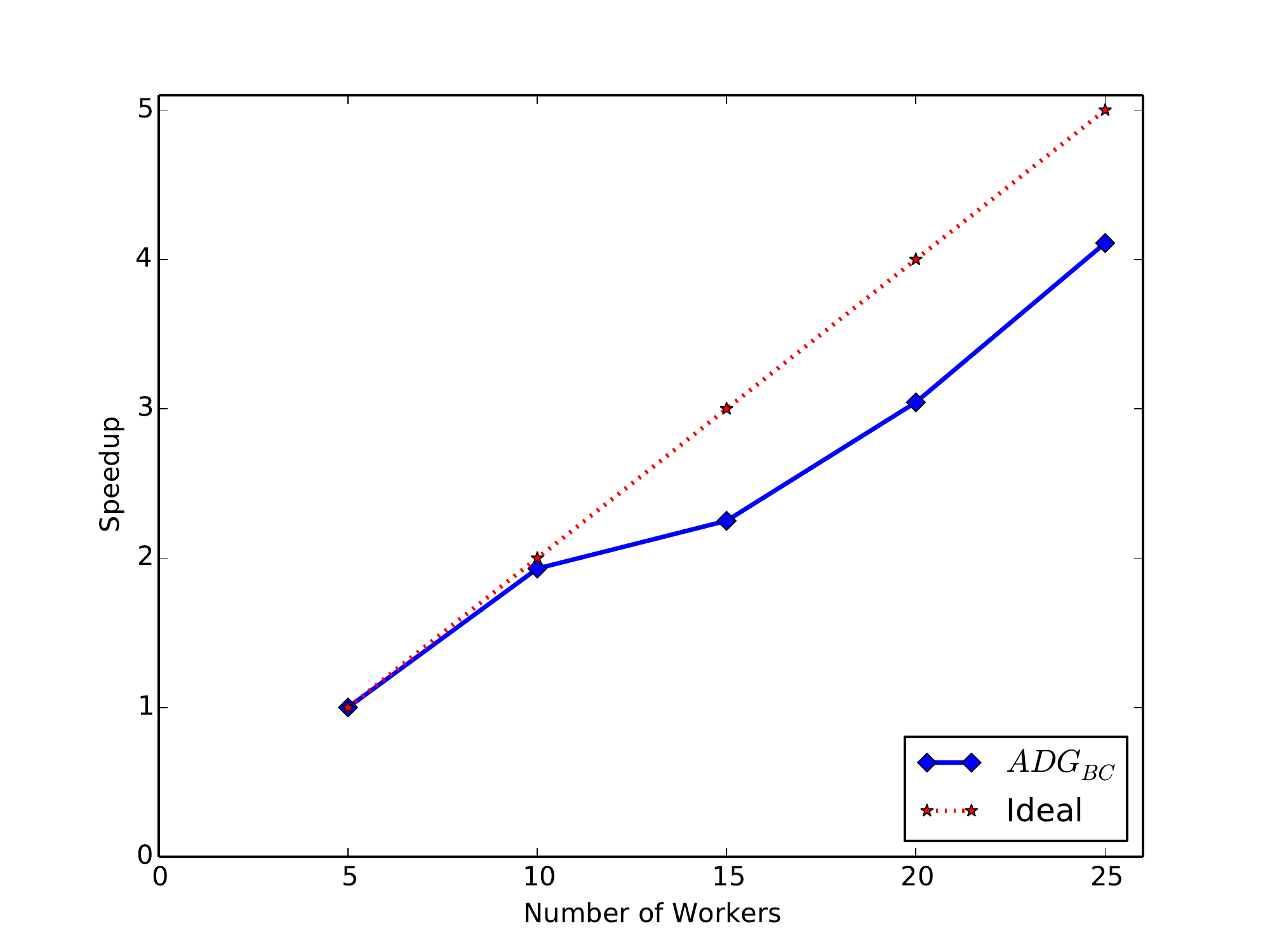}\\
      Loss Function Vs. Time (sec) & Speedup Vs. Number of Workers\\
  \end{tabular}%

  \caption{Convergence Speedup Result for \Epsilon\ Dataset}
  \label{fig:speedup}
\end{figure}

We also present the communication overhead incurred by each of the methods. The total communication cost for each algorithm is compared in terms of the total number of communication calls (send, receive, broadcast, gather), as well as the time spent in those calls. Since for $\SGrad$ and $\ASGrad$ methods the convergence is very slow near the tail, we compare the communication cost till the iteration when all methods achieve the same minimization of the objective function. Table \ref{tab:comparison_comm} shows the detailed results obtained for each algorithm on all datasets. It can be observed that the $\AGDBC$ incurs the minimum communication overhead as the number of communication between the machines is very low. Most of the calls shown for $\AGDBC$ are made during the first epoch where the gradients are synchronized. Whereas $\SGrad$ and $\ASGrad$ methods have to communicate large number of times in order to broadcast their local gradients to the master and receive the updated parameters from the master machine.

\subsubsection{Speedup Result with Increasing Number of Workers}

Finally, we evaluate the speedup in convergence (in terms of training loss and test accuracy) varying the number of workers from 5 to 25. Results shown in Figure \ref{fig:speedup} suggest that as the number of workers increases the $\AGDBC$ algorithm is able to achieve a near linear speedup, which is mainly due to the fact that, it relies on very low communication between the workers which is is also shown in Table \ref{tab:comparison_comm}. However, as the number of workers increases the performance of the algorithm slightly deteriorates.

\subsection{Experimental Results for Matrix Factorization}
\label{sec:Exp_MF}
We also conducted a number of experiments to empirically validate the proposed asynchronous framework on matrix factorization for recommendation where the recommendation matrix is split into $M$ rows as in Problem~\eqref{eqn:pb-split}.



\noindent\textbf{Datasets:} We performed experiments on  Movielens-10M (ML-10M)\footnote{http://grouplens.org/datasets/movielens/} and the Netflix Collection\footnote{http://www.netflixprize.com/} that are two popular corpora in collaborative filtering. 

\hspace{-5mm}\textbf{Baselines:}To validate the asynchronous distributed algorithm described in the previous section, we compare the following four strategies:
\begin{itemize} 
\item The proposed approach ${\AGDMF}$ (Section \ref{sec:MF}),
\item The asynchronous distributed ADMM approach (\texttt{AD-ADMM})  \cite{Tsung15},
\item Two distributed algorithms specifically proposed for matrix factorization {\ASGD} \cite{makari2015shared} and {\DSGD} \cite{gemulla2011large} (Section \ref{sec:MF}).
\end{itemize}

\hspace{-5mm}\textbf{Platform:} The distributed framework we considered was implemented using PySpark version 1.5.1. by connecting 7 servers with different computational power. 

\hspace{-5mm}\textbf{Hyper-Parameters:} Various free parameters of {\SGO} such as learning rate ($\gamma$), regularization parameter ($\lambda$) and number of latent factors ($K$) were set following \cite{chin2015learning}, \cite{yu2014distributed}. These values as well as the datasets characteristics  are listed in Table \ref{tab:dataset_properties}.

\begin{table}[h!]
\caption{Characteristics of Datasets used in our experiments. $|\mathcal U|$ and $|\mathcal I|$ denote respectively the number of users and items.} 
\label{tab:dataset_properties}
\tabcolsep=0.08cm
\centering
\resizebox{\textwidth}{!}{
  \begin{tabular}{ l c c c c c c c c}
  \hline
  Dataset & $|\mathcal U|$ & $|\mathcal I|$ & $\gamma$ & $\lambda$ & $K$ &  training size &  test size & sparsity\\
  \hline
  
    ML-10M & 71567 & 10681 & 0.005 & 0.05 & 100 &  9301274 & 698780 & 98.7 \%\\
  
    NetFlix (NF) & 480189 & 17770 & 0.005 & 0.05 & 40 &  99072112 & 1408395 & 99.8 \%\\

    NF-Subset & 28978 & 1821 & 0.005 & 0.05 & 40 &  3255352 & 100478 & 93.7 \%\\

  \hline
   
\end{tabular}
}
\end{table}

\setcounter{figure}{3}

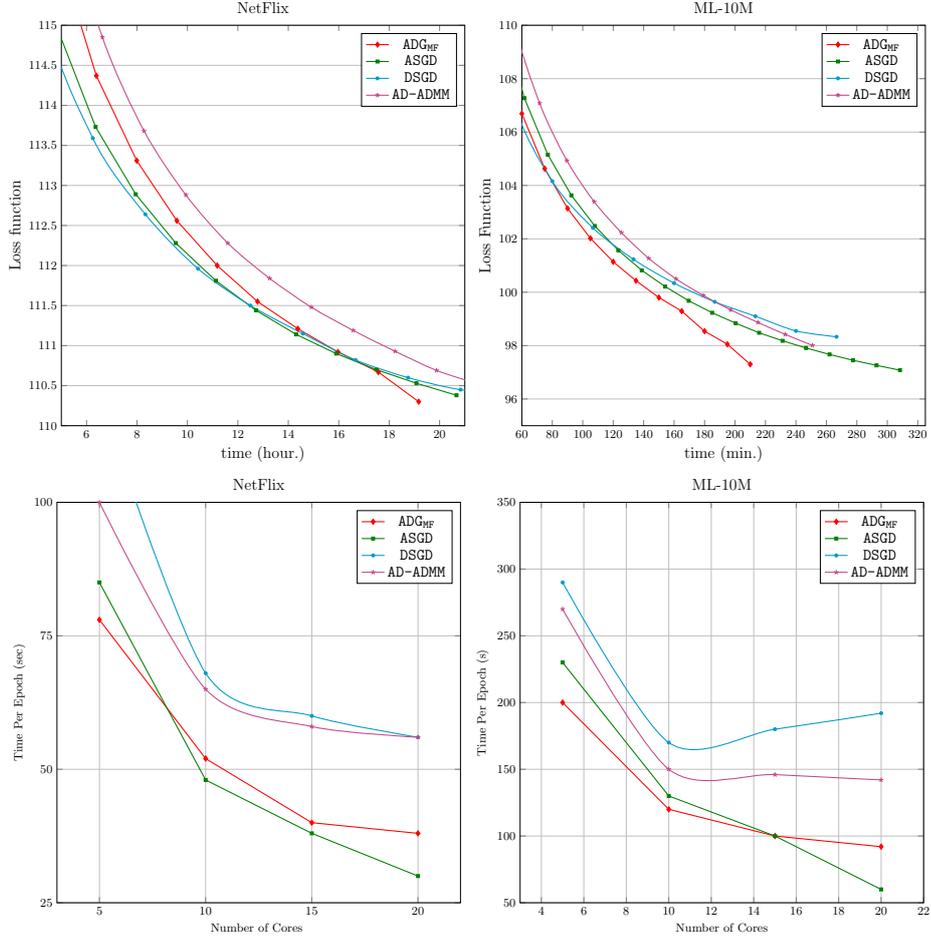
\begin{figure}[t!]
\begin{tabular}{c c}
  \begin{tikzpicture}[scale=.5]
  \begin{axis}[ 
	width=\textwidth,
	height=\textwidth,
	title=\text{\large NetFlix},
 ymajorgrids, 
yminorgrids,
   compat=newest, 
   xmin=5, xmax=21,
   ymin=110, 
   ymax=115,
   xlabel=\text{\large time (hour.)},
    ylabel=\text{\large Loss function},
    legend style={font=\tiny},
extra x tick style={grid=major, tick label style={yshift=0.55cm,rotate=90,anchor=west}},
  ]  
\node[draw] at (12,116) {some text};
\addplot [ 
 color=red,  
 line width=0.8pt, 
 mark size=1.9pt, 
 mark=diamond*,
 mark options={solid,fill=red,draw=red}, 
 ] 
coordinates {
(0.00,321.84)(1.60,129.39)(3.19,119.13)(4.79,116.01)(6.39,114.37)(7.99,113.31)(9.58,112.56)(11.18,112.00)(12.78,111.55)(14.38,111.21)(15.97,110.92)(17.57,110.67)(19.17,110.30)
};
\addlegendentry{\large $\AGDMF$}

\addplot [ 
 color=green!50!black,  
 line width=0.8pt, 
 mark size=1.1pt, 
 mark=square*,
 mark options={solid,fill=green!50!black,draw=green!50!black}, 
 ] 
coordinates {

(0.00,322.18)(1.59,123.99)(3.18,117.48)(4.77,115.02)(6.36,113.73)(7.95,112.89)(9.54,112.28)(11.13,111.81)(12.72,111.44)(14.31,111.14)(15.90,110.90)(17.49,110.70)(19.08,110.53)(20.67,110.38)

};
\addlegendentry{\large \ASGD}

\addplot [ 
 color=cyan!80!black,  
 smooth,
 line width=0.8pt, 
 mark size=1.1pt, 
 mark=*,
 mark options={solid,fill=cyan!80!black,draw=cyan!80!black}, 
 ]  coordinates {
(0.00,284.20)(2.08,120.02)(4.17,115.40)(6.25,113.59)(8.33,112.64)(10.42,111.96)(12.50,111.50)(14.58,111.15)(16.67,110.82)(18.75,110.60)(20.83,110.45)(22.92,110.41)

};
\addlegendentry{\large \DSGD}

\addplot [ 
 color=magenta!80!black,  
 smooth,
 line width=0.8pt, 
 mark size=1.8pt, 
 mark=star,
 ]  coordinates {
(1.66,133.73)(3.31,120.46)(4.97,116.70)(6.63,114.85)(8.28,113.68)(9.94,112.88)(11.60,112.28)(13.26,111.84)(14.91,111.48)(16.57,111.19)(18.23,110.93)(19.88,110.69)(21.54,110.52)

};
\addlegendentry{\large \texttt{AD-ADMM}}
{font=\fontsize{4}{5}\selectfont}

\end{axis}
\end{tikzpicture} 
&
  \begin{tikzpicture}[scale=.5]
  \begin{axis}[ 
	width=\textwidth,
	height=\textwidth,
   compat=newest, 
   ymajorgrids,
   xmin=60, xmax=325,
     ymin=95, 
     ymax=110,
     title=\text{\large ML-10M},
   xlabel=\text{\large time (min.)},
    ylabel=\text{\large Loss Function},
    legend style={font=\tiny},
extra x tick style={grid=major, tick label style={yshift=1.5cm,rotate=90,anchor=west}},
  ]  

\addplot [ 
 color=red,  
 line width=0.8pt, 
 mark size=1.9pt, 
 mark=diamond*,
 mark options={solid,fill=red,draw=red}, 
 ]  coordinates {
 
(0,621.97)(15,135.34)(30,115.73)(45,109.89)(60,106.69)(75,104.63)(90,103.14)(105,102.02)(120,101.14)(135,100.43)(150,99.8)(165,99.29)(180,98.54)(195,98.05)(210,97.3)

};
\addlegendentry{\large $\AGDMF$}

\addplot [ 
 color=green!50!black,  
 line width=0.8pt, 
 mark size=1.1pt, 
 mark=square*,
 mark options={solid,fill=green!50!black,draw=green!50!black}, 
 ] coordinates {
(0.00, 627.18)(15.42, 131.33)(30.83, 116.2)(46.25, 110.49)(61.67, 107.27)(77.08, 105.15)(92.50, 103.63)(107.92, 102.48)(123.33, 101.57)(138.75, 100.82)(154.17, 100.21)(169.58, 99.68)(185.00, 99.23)(200.42, 98.84)(215.83, 98.48)(231.25, 98.18)(246.67, 97.91)(262.08, 97.67)(277.50, 97.45)(292.92, 97.26)(308.33, 97.08)
};
\addlegendentry{\large \ASGD}

\addplot [ 
 color=cyan!80!black,  
 smooth,
 line width=0.8pt, 
 mark size=1.1pt, 
 mark=*,
 mark options={solid,fill=cyan!80!black,draw=cyan!80!black}, 
 ] coordinates {
(0.00,517.82)(26.67,115.25)(53.33,107.39)(80.00,104.16)(106.67,102.41)(133.33,101.23)(160.00,100.34)(186.67,99.64)(213.33,99.1)(240.00,98.55)(266.67,98.33)};
\addlegendentry{\large \DSGD}

\addplot [ 
 color=magenta!80!black,  
 smooth,
 line width=0.8pt, 
 mark size=1.8pt, 
 mark=star,
 ]  coordinates {
(0.00,621.79)(17.92,136.25)(35.83,116.48)(53.75,110.38)(71.67,107.08)(89.58,104.93)(107.50,103.39)(125.42,102.23)(143.33,101.27)(161.25,100.5)(179.17,99.88)(197.08,99.34)(215.00,98.87)(232.92,98.42)(250.83,98.01)};
\addlegendentry{\large \texttt{AD-ADMM}}

{font=\fontsize{4}{5}\selectfont}

\end{axis}

\end{tikzpicture}
 

\\


  \begin{tikzpicture}[scale=.5]
  \begin{axis}[ 
	width=\textwidth,
	height=\textwidth,
 xmajorgrids, 
 ymajorgrids, 
yminorgrids,
   compat=newest, 
   xmin=3, xmax=22,
   ymin=25, 
   ymax=100,
   title=\text{\large NetFlix},
    xtick={5.,10, 15, 20},
    ytick={25, 50, 75, 100},
    xlabel=\text{\small Number of Cores },
    ylabel=\text{\small Time Per Epoch (sec)},
    legend style={font=\tiny},
  ]  

\addplot [ 
 color=red,  
 line width=0.8pt, 
 mark size=1.9pt, 
 mark=diamond*,
 mark options={solid,fill=red,draw=red}, 
 ] 
coordinates {
(5, 78) (10, 52) (15, 40) (20, 38)
};
\addlegendentry{\large $\AGDMF$}

\addplot [ 
 color=green!50!black,  
 line width=0.8pt, 
 mark size=1.1pt, 
 mark=square*,
 mark options={solid,fill=green!50!black,draw=green!50!black}, 
 ] 
coordinates {
(5, 85) (10, 48)  (15, 38) (20, 30)

};
\addlegendentry{\large \ASGD}

\addplot [ 
 color=cyan!80!black,  
 smooth,
 line width=0.8pt, 
 mark size=1.1pt, 
 mark=*,
 mark options={solid,fill=cyan!80!black,draw=cyan!80!black}, 
 ]  coordinates {
(5, 120) (10, 68) (15, 60) (20, 56)
};
\addlegendentry{\large \DSGD}

\addplot [ 
 color=magenta!80!black,  
 smooth,
 line width=0.8pt, 
 mark size=1.8pt, 
 mark=star,
 ]  coordinates {
(5,100) (10, 65) (15, 58) (20, 56)
};
\addlegendentry{\large \texttt{AD-ADMM}}
{font=\fontsize{4}{5}\selectfont}

\end{axis}
\end{tikzpicture} &
  \begin{tikzpicture}[scale=.5]
  \begin{axis}[ 
	width=\textwidth,
	height=\textwidth,
   compat=newest, 
   xmajorgrids,
   ymajorgrids,
   xmin=3, xmax=22,
     ymin=50, 
     title=\text{\large ML-10M},
     ymax=350,
   xlabel=\text{\small Number of Cores},
    ylabel=\text{\small Time Per Epoch (s)},
    legend style={font=\tiny},
  ]  

\addplot [ 
 color=red,  
 line width=0.8pt, 
 mark size=1.9pt, 
 mark=diamond*,
 mark options={solid,fill=red,draw=red}, 
 ]  coordinates {
(5, 200)(10, 120)(15, 100)(20, 92)
};
\addlegendentry{\large $\AGDMF$}

\addplot [ 
 color=green!50!black,  
 line width=0.8pt, 
 mark size=1.1pt, 
 mark=square*,
 mark options={solid,fill=green!50!black,draw=green!50!black}, 
 ] coordinates {
(5, 230) (10, 130) (15, 100) (20, 60)
};
\addlegendentry{\large \ASGD}

\addplot [ 
 color=cyan!80!black,  
 smooth,
 line width=0.8pt, 
 mark size=1.1pt, 
 mark=*,
 mark options={solid,fill=cyan!80!black,draw=cyan!80!black}, 
 ] coordinates {
 (5, 290) (10, 170) (15, 180) (20, 192)
};
\addlegendentry{\large \DSGD}

\addplot [ 
 color=magenta!80!black,  
 smooth,
 line width=0.8pt, 
 mark size=1.8pt, 
 mark=star,
 ]  coordinates {
 (5, 270) (10, 150) (15, 146) (20, 142)
};
\addlegendentry{\large \texttt{AD-ADMM}}

{font=\fontsize{4}{5}\selectfont}

\end{axis}
\end{tikzpicture}
 \end{tabular}
\caption{\textbf{Top}: Test RMSE curves with respect to time for {$\AGDMF$}, \texttt{AD-ADMM}, {\ASGD},  and {\DSGD} on NetFlix (left), and ML-10M (right) Datasets.
\textbf{Bottom}: Total Convergence Time Vs. Number of Cores curves for ${\AGDMF}$, {\ASGD}, {\DSGD} and \texttt{AD-ADMM} on the NetFlix (left), and ML-10M (right)  Datasets.}
\label{fig:RMSE_Curve}

\end{figure}
\subsubsection{Evaluation of Convergence Time}

We begin our experiments by comparing the evolution of the loss function of Eq.~\eqref{eqn:pb} with respect to time until convergence. The convergence points are shown as names of the algorithms vertically (we stopped ASGD after 20 hours on the NF dataset). Figure \ref{fig:RMSE_Curve} (top) depicts this evolution for ML-10M and NF datasets using 10 and 15 cores respectively. Synchronization based approaches ({\ASGD} and {\DSGD}) aggregate all the information at each epoch and thus begin to converge more sharply at the beginning. However, with these approaches, when the fastest machines finish their computations, they have to wait for slower machines; thus, they require much more time to converge than the asynchronous methods (\texttt{AD-ADMM} and ${\AGDMF}$). 
Finally, it comes out that ${\AGDMF}$ converges faster than the other algorithms on both datasets. This is mainly due to the fact that ${\AGDMF}$ does not obey to any delay mechanism as in \texttt{AD-ADMM} for instance.

\subsubsection{Computation and Communication Trade-off }

We performed another set of experiments aimed at measuring the effect of number of cores on performance of the proposed approach and the baselines. Figure \ref{fig:RMSE_Curve} (bottom) depicts this effect by showing the evolution of time per epoch of the SGD method used in ${\AGDMF}$, \texttt{ASGD}, \texttt{DSGD} and \texttt{AD-ADMM} with respect to increasing number of machines. From these experiments, it comes out that for all approaches the time per epoch of the $\SGO$ method decreases as the number of machines increases. 

But after a certain number of machines (10 in both experiments), the time per epoch of some approaches begin to be affected as the communication cost takes over the computation time. The approach that is the most affected by this is \texttt{DSGD}, as synchronizations in this case are done after each sub-epoch. We can also see that even though the per epoch speedup is best for \texttt{ASGD}, it requires a much higher number of epochs to converge as compared to ${\AGDMF}$ and \texttt{DSGD}. 

%% file: conclusion.tex
\section{Conclusion}

In this paper we proposed a novel asynchronous distributed framework for the minimization of general smooth objective functions that write as a sum of instantaneous loss functions, where parameters are exchanged rather than gradients which is the case for almost the majority of distributed learning algorithms. We proved the consistency of this approach when the elementary operation at each node is a gradient descent. Then, we built upon this framework to propose two asynchronous distributed algorithms for: matrix factorization for recommender systems and large scale binary classification. Then we empirically validated effectiveness of the two proposed algorithms in corresponding application domains. As a perspective, we aim at extending this work by considering additional proximal operations in order to deal with non-smooth convex functions as well as broad regularization terms.